# A Computational Treatment of HPSG Lexical Rules as Covariation in Lexical Entries

Walt Detmar Meurers, Guido Minnen*

SFB 340, Kleine Wilhelmstr. 113, 72074 Tübingen, Germany
E-mail: {dm,minnen}@sfs.nphil.uni-tuebingen.de

## Abstract

We describe a compiler which translates a set of HPSG lexical rules and their interaction into definite relations used to constrain lexical entries. The compiler ensures automatic transfer of properties unchanged by a lexical rule. Thus an operational semantics for the full lexical rule mechanism as used in HPSG linguistics is provided. Program transformation techniques are used to advance the resulting encoding. The final output constitutes a computational counterpart of the linguistic generalizations captured by lexical rules and allows "on the fly" application.

## 1 Introduction

In the paradigm of HPSG, lexical rules (henceforth LR) have become one of the key mechanisms used in current analysis. Among logicians and computational linguists, LRs have been far less popular. The intuitive idea behind LRs is based on notions such as matching, copying, and automatic transfer of the properties unchanged by a LR, which are not easily integrated in the logic setup of HPSG. Even though some studies of various formal and computational aspects of HPSG LRs exist (e.g., [FPW85], [Fli87], [Pol93], and [Gei94]), so far no proposal captures the full semantics intended for the LR mechanism. For computational linguists, LRs cause additional problems since they are a completely unrestricted mechanism for enlarging the lexicon. In the general case it is impossible to decide beforehand which LR derives a lexical entry (henceforth LE) meeting a certain requirement. The usual computational treatment of LRs therefore computes all entries resulting from LR applications at compile time (e.g., [Car92]). The generalizations which were captured by LRs are lost for computation. Moreover, such a treatment cannot be used for the increasing number of HPSG theories which propose LRs that result in an infinite lexicon.

The research reported here was sponsored by Teilprojekt B4 "From Constraints to Rules: Efficient Compilation of HPSG Grammars" of SFB 340 "Sprachtheoretische Grundlagen für die Computerlinguistik" of the Deutsche Forschungsgemeinschaft. The authors wish to thank Thilo Götz and Dale Gerdemann, Erhard Hinrichs, Paul King, Dieter Martini, Bill Rounds as well as three anonymous reviewers for comments and discussion. Of course, the authors are responsible for all remaining errors.

*The authors are listed alphabetically. URL: http://www.sfs.nphil.uni-tuebingen.de/sfb/

We investigate a new computational treatment of LRs which, instead of expressing relations between LEs, encodes possible variations resulting from LR application inside of the entries. A similar proposal has been made in [vNB94]. Contrary to their use of this method for hand encoding one LR, we interpret LRs in general as systematic covariation in LEs. I.e., we establish a formal link between the LR mechanism and definite relations encoding covariation in LEs. The interaction of a set of LRs and the transfer of all properties unchanged by a LR are automatically deduced from the set of LRs provided.

We developed a compiler that translates a set of LRs and their interaction into definite relations constraining LEs. We show that the definite relations produced by the compiler can be refined by program transformation techniques to increase run-time efficiency without losing an independent representation of LRs. In addition the compiler adapts the lexicon such that LEs directly bear all specifications not changed by the LRs. This permits delayed evaluation of lexical covariation, i.e., "on the fly" application of LRs, which avoids expanding out the lexicon.

The conception of LRs underlying the research presented here makes it possible to deal with the full LR mechanism within the feature logic for HPSG proposed in [Kin89] and [Kin94]. The reader is referred to [Meu] for the formal semantics of LRs upon which our computational treatment is based. Due to this theoretical foundation, the computational treatment of LRs proposed can be seen as an extension to the principled method discussed in [GM95] for encoding the main building block of HPSG grammars – the implicative constraints – as a logic program.

The structure of the paper is as follows. In section 2 we show how LRs and their interaction can be expressed as systematic covariation in LEs and how abstract lexicon expansion is used to produce definite relations encoding LRs and their interaction. Subsequently we focus on an improvement of abstract lexicon expansion by means of program transformation techniques (section 3). Section 4 presents an adaption of the lexicon necessary for "on the fly" application of LRs.

# 2 Lexical Covariation: Encoding Lexical Rules and their Interaction as Definite Relations

The treatment of LRs we investigate in this paper at first sight differs significantly from the conventional view of LRs as relations between LEs. We express the application of a set of LRs as definite relations encoding systematic covariation in *base lexical entries*. I.e., inside of the LEs which are supposed to feed one or more LRs, we express the LEs which can be derived from them by means of different solutions to calls to definite relations.

By encoding LR application inside of each LE, generalizations over classes of LEs seem to be lost in our approach. This is not true though, since we use a call to the same definite relation in all of a natural class[1] of LEs. This way definite relations capture the systematic covariation in LEs belonging to a particular natural class. We are able to maintain an independent definite clause representation of the LRs specified by the linguist to express generalizations over the lexicon.

[1] In section 2.3 we show how the compiler detects such natural classes.

In the following, we describe four compilation steps which translate a set of LRs as specified by the linguist and their interaction into definite relations to constrain LEs.

## 2.1 Lexical Rules as Definite Relations and the Automatic Property Transfer

We start by translating each LR into a definite clause predicate, called the *lexical rule predicate*. The first argument of a LR predicate corresponds to the in-specification of the LR and the second argument to its out-specification.

Assume the following signature (the type hierarchy and the appropriateness conditions), which will be used throughout the paper:[2]

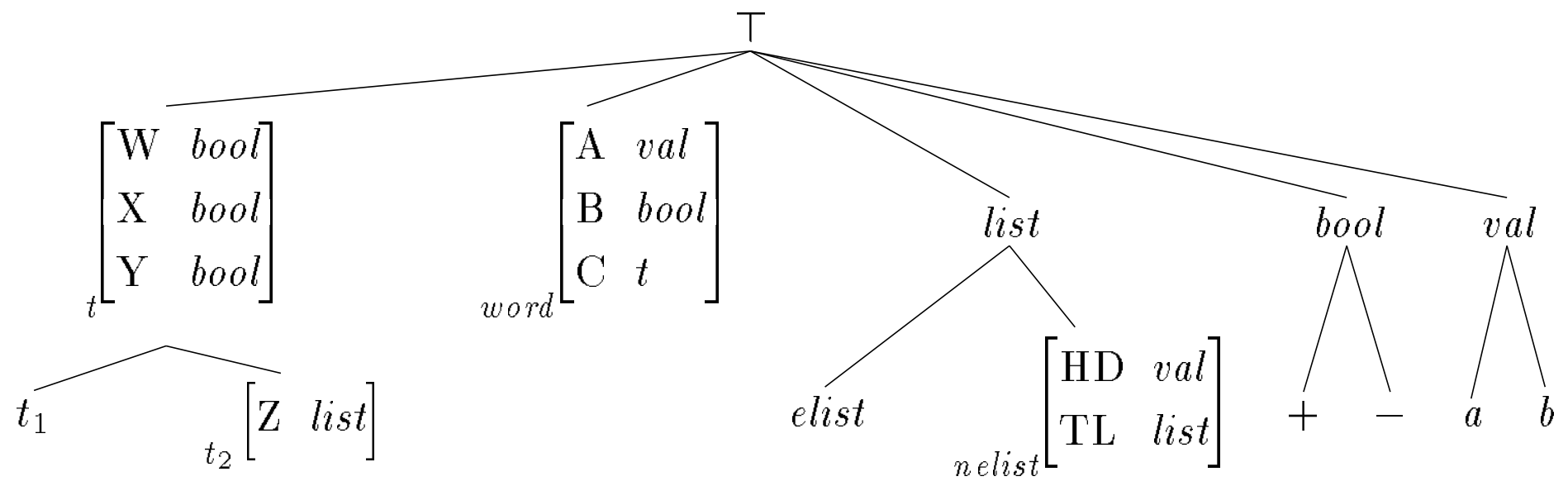

Figure 1: An example signature

Suppose the linguist specifies the following LR:

$$\begin{bmatrix} \text{B} & - \\ \text{C} & {}_{t}\begin{bmatrix}\text{Y} & -\end{bmatrix}\end{bmatrix} \mapsto \begin{bmatrix} \text{A} & b \\ \text{C} & {}_{t}\begin{bmatrix}\text{X} & + \\ \text{Y} & +\end{bmatrix}\end{bmatrix}$$

Figure 2: LR 1 as specified by the linguist

This LR applies to LEs which unify with the in-specification, i.e., LEs which specify B and Y as $-$. The derived LE licenses *word*-objects with $+$ as the value of X and Y, and $b$ as that of A. The translation of the LR into a predicate is trivial:

$$\text{lex_rule_1}(\begin{bmatrix} \text{B} & - \\ \text{C} & {}_{t}\begin{bmatrix}\text{Y} & -\end{bmatrix}\end{bmatrix}, \begin{bmatrix} \text{A} & b \\ \text{C} & {}_{t}\begin{bmatrix}\text{X} & + \\ \text{Y} & +\end{bmatrix}\end{bmatrix})$$

Figure 3: Definite clause representation of LR 1

Though this predicate represents what the linguist specified, it does not accomplish exactly what she/he intended. Features specified in a LE unifying with the in-specification of the LR which are not specified differently in the out-specification of the LR are intended to receive the same value on the derived entry as on the base entry, i.e., additional path equations between the in- and the out-specification of the LRs have to be ensured. We will call this *automatic property transfer*.

[2]Space limitations force us to use rather abstract LRs in the examples of this paper.

The detection of which additional path equations are intended by the linguist crucially depends on the closed world interpretation of the type hierarchy assumed in HPSG. The closed world interpretation makes it possible to determine which kind of (by ontological status fully specific)[3] lexical objects may undergo the rule. Since a type can always be replaced by a disjunction of its minimal subtypes, on the basis of the signature we can determine which paths the linguist left unspecified in the out-specification of the LR. This allows us to "fill in" path equalities between the in- and the out-specification of the LR to make sure that the values of all features which do not get specified differently in the out-specification get transferred.

However, deriving automatic property transfer for a LR can be problematic because the in- and out-specification of a LR are usually less specific than the base LEs which license the input words. In fact, the LR 1 of our example applies to LEs with $t_1$ as their C value and to those having $t_2$ as their C value. With respect to property transfer this means that there can be LEs such as

$$_{word}\begin{bmatrix} \text{C} & t_1 \end{bmatrix}$$

for which we need to make sure that $t_1$ as the value of C gets transferred. However, the type information $t_1$, which is more specific than that given in the output of the LR, can only be specified on the out-specification of the LR if the specification of C is transferred as a whole (via structure sharing of the value of C). This is not possible since the values of X and Y are specified in the out-specification of the LR. In more general terms the problem is that there is no notion of sharing just the type of an object. However, not only such typing information, but also certain feature values can get lost. The subtypes of $t$ have different appropriate features, the values of which possibly have to be preserved. In case the LE has $t_2$ as the value of C, we need to ensure that the value of the feature Z is transferred properly.

To ensure that no information is lost as a result of applying a LR, it seems to be necessary to split up the LR and make each instance more specific. In the above example[4] this would result in two LRs: one for words with $t_1$ as their C value and one for those with $t_2$ as their C value. In the latter case we can also take care of transferring the value of Z. However, as discussed in [Meu94], making several instances of LRs can be avoided. Instead, the disjunctive possibilities introduced by property transfer can be pushed inside of a LR. This is accomplished by having each LR predicate call a so-called *transfer predicate* which can have multiple defining clauses. So for the LR 1, property transfer is taken care of by extending the predicate in figure 3 with a call to a transfer predicate in the following way:

$$\text{lex_rule_1}(\boxed{1}\begin{bmatrix} \text{B} & - \\ \text{C} & {}_{t}\begin{bmatrix} \text{Y} & - \end{bmatrix} \end{bmatrix}, \boxed{2}\begin{bmatrix} \text{A} & b \\ \text{C} & {}_{t}\begin{bmatrix} \text{X} & + \\ \text{Y} & + \end{bmatrix} \end{bmatrix}) \text{:- transfer_1}(\boxed{1},\boxed{2}).$$

[3] A formalization of the ontology for HPSG assumed in [PS94] is provided in [Kin89] and [Kin94].

[4] A linguistic example based on the signature given in [PS94] would be a LR deriving predicative signs from non-predicative ones, i.e. changing the PRD value of substantive signs from $-$ to $+$ much like the LR for NPs given in [PS94, p. 360, fn. 20]. In such a PRD-LR (which we only mention as an example and not as a linguistic proposal) the subtype of the *head*-object undergoing the rule as well as the value of the features only appropriate for the subtypes of *substantive* is either lost or a separate rule for each of the subtypes has to be specified.

Figure 4: Lexical rule predicate representing lexical rule 1

On the basis of the LR specification and the signature, the compiler deduces the transfer predicates without requiring additional specifications by the linguist. The transfer predicate for LR 1 is defined by the following two clauses:

$$\text{transfer_1}(\begin{bmatrix} \text{B} & \boxed{1} \\ \text{C} & {}_{t_1}\begin{bmatrix} \text{W} & \boxed{2} \end{bmatrix} \end{bmatrix}, \begin{bmatrix} \text{B} & \boxed{1} \\ \text{C} & {}_{t_1}\begin{bmatrix} \text{W} & \boxed{2} \end{bmatrix} \end{bmatrix}). \qquad \text{transfer_1}(\begin{bmatrix} \text{B} & \boxed{1} \\ \text{C} & {}_{t_2}\begin{bmatrix} \text{W} & \boxed{2} \\ \text{Z} & \boxed{3} \end{bmatrix} \end{bmatrix}, \begin{bmatrix} \text{B} & \boxed{1} \\ \text{C} & {}_{t_2}\begin{bmatrix} \text{W} & \boxed{2} \\ \text{Z} & \boxed{3} \end{bmatrix} \end{bmatrix}).$$

Figure 5: Definition of the transfer predicate for lexical rule 1

The first case applies to LEs in which C is specified as $t_1$. We have to ensure that the value of the feature W is transferred. In the second case, when feature C has $t_2$ as its value, this does not suffice as we additionally have to ensure that Z gets transferred. Note that neither clause of the transfer predicate needs to specify the features A, X and Y as these features are changed by *lex_rule_1*. Furthermore, filling in features of the structure below Z is unnecessary as the value of Z is structure shared as a whole.

## 2.2 Determining Global Lexical Rule Interaction

In the second compilation step, we use the definite clause representation of a set of LRs, i.e., the LR and the transfer predicates, to compute a finite state automaton (henceforth FSA) representing how the LRs interact (irrespective of the LEs). In general, any LR can apply to the output of another LR, which is sometimes referred to as *free application*. When looking at a specific set of LRs though, it is possible to determine which LRs can possibly follow which LRs in that grammar. The set of follow-relationships is obtained by testing which in-specifications unify with which out-specifications.[5] Using this information, it is possible to avoid trying LR applications at run-time that are bound to fail.

To illustrate this first and the following steps determining global LR interaction, let us add three more LRs to the one discussed in 2.1. Figure 6 shows the full set of four

[5] For the computation of the follow-relationships, the specifications of the transfer predicates are taken into account. In case the transfer relation called by a LR has several defining clauses, the generalization of the transfer possibilities is used.

LRs as specified by the linguist.

Rule 1: $\begin{bmatrix} \text{B} & - \\ \text{C} & \begin{bmatrix} \text{Y} & - \end{bmatrix} \end{bmatrix} \mapsto \begin{bmatrix} \text{A} & b \\ \text{C} & \begin{bmatrix} \text{X} & + \\ \text{Y} & + \end{bmatrix} \end{bmatrix}$

Rule 2: $\begin{bmatrix} \text{A} & b \\ \text{B} & - \\ \text{C} & \begin{bmatrix} \text{W} & - \end{bmatrix} \end{bmatrix} \mapsto \begin{bmatrix} \text{C} & \begin{bmatrix} \text{W} & + \end{bmatrix} \end{bmatrix}$

Rule 3: $\begin{bmatrix} \text{C}_{t_2} & \begin{bmatrix} \text{W} & + \\ \text{X} & + \\ \text{Z}\,|\,\text{TL} & \boxed{1} \end{bmatrix} \end{bmatrix} \mapsto \begin{bmatrix} \text{C}_{t_2} & \begin{bmatrix} \text{Y} & + \\ \text{Z} & \boxed{1} \end{bmatrix} \end{bmatrix}$

Rule 4: $\begin{bmatrix} \text{B} & - \\ \text{C}_{t_2} & \begin{bmatrix} \text{W} & + \\ \text{X} & + \\ \text{Z} & \langle\rangle \end{bmatrix} \end{bmatrix} \mapsto \begin{bmatrix} \text{B} & + \\ \text{C}_{t_2} & \begin{bmatrix} \text{X} & - \end{bmatrix} \end{bmatrix}$

Figure 6: A set of four lexical rules as specified by the linguist

The following figure shows the definite clause representations of LRs 2, 3 and 4 and the transfer predicates derived for them. The definite clauses representing LR 1 and its transfer were already given in figures 4 and 5. The follow-relation obtained for the set of four LRs is shown in the figure 8.

$\text{lex_rule_2}(\boxed{1}\begin{bmatrix} \text{A} & b \\ \text{B} & - \\ \text{C} & \begin{bmatrix} \text{W} & - \end{bmatrix} \end{bmatrix}, \boxed{2}\begin{bmatrix} \text{C} & \begin{bmatrix} \text{W} & + \end{bmatrix} \end{bmatrix})\text{:- transfer_2}(\boxed{1},\boxed{2}).$

$\text{lex_rule_3}(\boxed{1}\begin{bmatrix} \text{C}_{t_2} & \begin{bmatrix} \text{W} & + \\ \text{X} & + \\ \text{Z}\,|\,\text{TL} & \boxed{3} \end{bmatrix} \end{bmatrix}, \boxed{2}\begin{bmatrix} \text{C}_{t_2} & \begin{bmatrix} \text{Y} & + \\ \text{Z} & \boxed{3} \end{bmatrix} \end{bmatrix})\text{:- transfer_3}(\boxed{1},\boxed{2}).$

$\text{lex_rule_4}(\boxed{1}\begin{bmatrix} \text{B} & - \\ \text{C}_{t_2} & \begin{bmatrix} \text{W} & + \\ \text{X} & + \\ \text{Z} & \langle\rangle \end{bmatrix} \end{bmatrix}, \boxed{2}\begin{bmatrix} \text{B} & + \\ \text{C}_{t_2} & \begin{bmatrix} \text{X} & - \end{bmatrix} \end{bmatrix})\text{:- transfer_4}(\boxed{1},\boxed{2}).$

$\text{transfer_2}(\begin{bmatrix} \text{A} & \boxed{1} \\ \text{B} & \boxed{2} \\ \text{C}_{t_1} & \begin{bmatrix} \text{X} & \boxed{3} \\ \text{Y} & \boxed{4} \end{bmatrix} \end{bmatrix}, \begin{bmatrix} \text{A} & \boxed{1} \\ \text{B} & \boxed{2} \\ \text{C}_{t_1} & \begin{bmatrix} \text{X} & \boxed{3} \\ \text{Y} & \boxed{4} \end{bmatrix} \end{bmatrix}).$

$\text{transfer_2}(\begin{bmatrix} \text{A} & \boxed{1} \\ \text{B} & \boxed{2} \\ \text{C}_{t_2} & \begin{bmatrix} \text{X} & \boxed{3} \\ \text{Y} & \boxed{4} \\ \text{Z} & \boxed{5} \end{bmatrix} \end{bmatrix}, \begin{bmatrix} \text{A} & \boxed{1} \\ \text{B} & \boxed{2} \\ \text{C}_{t_2} & \begin{bmatrix} \text{X} & \boxed{3} \\ \text{Y} & \boxed{4} \\ \text{Z} & \boxed{5} \end{bmatrix} \end{bmatrix}).$

$\text{transfer_3}(\begin{bmatrix} \text{A} & \boxed{1} \\ \text{B} & \boxed{2} \\ \text{C}_{t_2} & \begin{bmatrix} \text{W} & \boxed{3} \\ \text{X} & \boxed{4} \end{bmatrix} \end{bmatrix}, \begin{bmatrix} \text{A} & \boxed{1} \\ \text{B} & \boxed{2} \\ \text{C}_{t_2} & \begin{bmatrix} \text{W} & \boxed{3} \\ \text{X} & \boxed{4} \end{bmatrix} \end{bmatrix}).$

$\text{transfer_4}(\begin{bmatrix} \text{A} & \boxed{1} \\ \text{C}_{t_2} & \begin{bmatrix} \text{W} & \boxed{2} \\ \text{Y} & \boxed{3} \\ \text{Z} & \boxed{4} \end{bmatrix} \end{bmatrix}, \begin{bmatrix} \text{A} & \boxed{1} \\ \text{C}_{t_2} & \begin{bmatrix} \text{W} & \boxed{2} \\ \text{Y} & \boxed{3} \\ \text{Z} & \boxed{4} \end{bmatrix} \end{bmatrix}).$

Figure 7: The definite clause encoding of lexical rules 2, 3, and 4

follow(1, [2, 3, 4]). follow(2, [1, 3, 4]). follow(3, [3, 4]). follow(4, []).

Figure 8: The follow-relation for the four lexical rules of the example

Once the follow-relation has been obtained, it can be used to construct an automaton that represents which LR can be applied after which *sequence* of LRs. Special care has to be taken in case the same LR can apply several times in a sequence. To obtain a *finite* automaton, such a repetition is encoded as a transition cycling back to a state in the LR sequence preceding it. In order to be able to (in the following steps) remove a transition representing a certain LR application in one sequence without eliminating the LR application from other sequences, every transition except for the ones introducing cycles leads to a new state. Otherwise we would obtain an automaton consisting of a single state with a cycle from/into this state for each of the LRs.

The FSA below is constructed on the basis of the follow-relation of figure 8. The state annotated with an angle bracket represents the initial state. All states (including the initial state) are final states. The labels of the transitions from one state to another are the LR predicate indices, i.e., the LR names constitute the alphabet of the FSA.

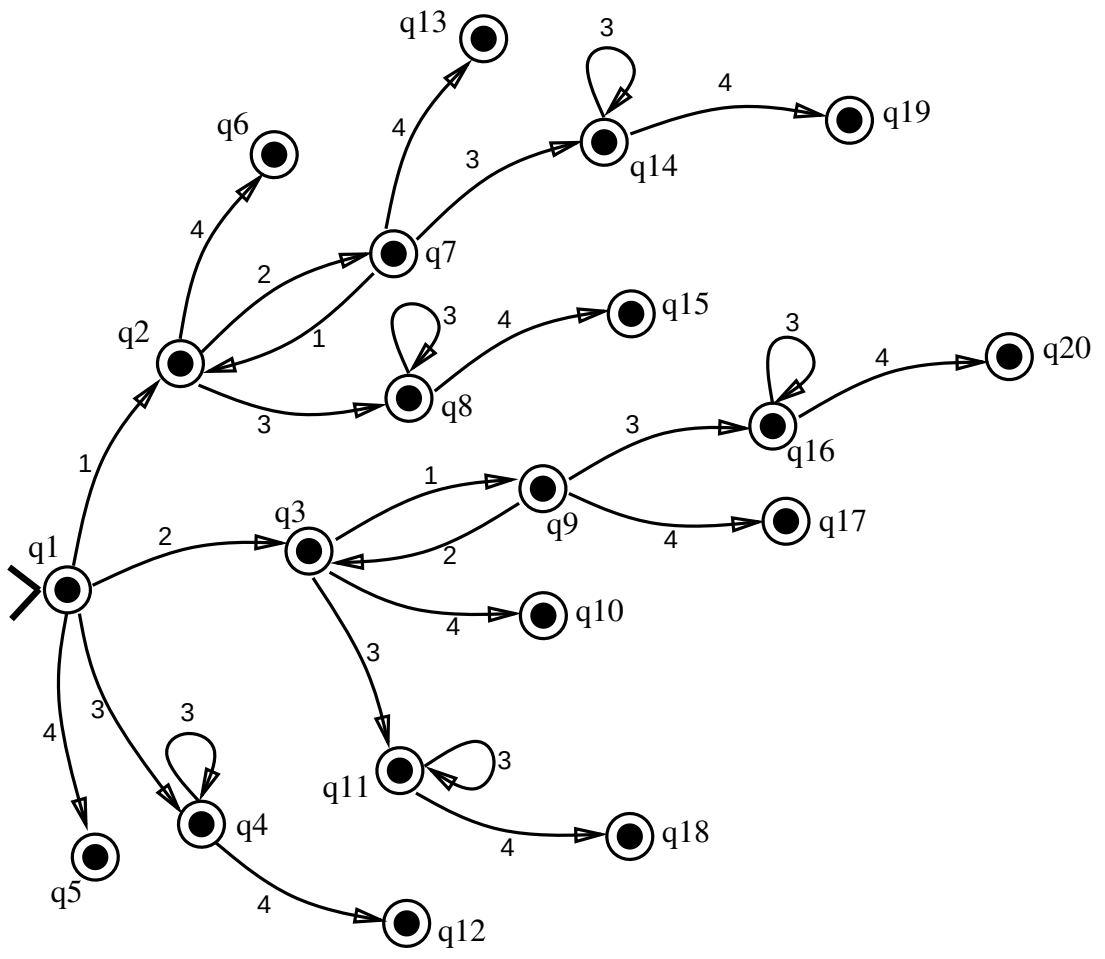

Figure 9: Finite state automaton representing global lexical rule interaction

Once such a FSA representing global LR interaction has been obtained, it can be used as the backbone of a definite clause encoding of LRs and their interaction (cf. 2.4).

Compared to free application, the FSA in figure 9 limits the choice of LRs possibly applying at a certain point. However, there are still several places where the choices can be further reduced. One possible reduction of the above automaton consists of taking into account the *propagation of specifications* along each possible path through the automaton. This corresponds to actually unifying the out-specification of a LR with the in-specification of the following LR along each path in the automaton, instead of merely testing for unifiability which we did to obtain the follow-relation. As a result of unifying the out-specification of a LR in a path of the FSA with the in-specification of the following LR, because of the structure sharing between the second LR's in- and out-specification (stemming from the LR and its property transfer), the out-specification of the second rule can become more specific. This makes it possible to eliminate some of the transitions which seem to be possible when judging on the basis of the follow-relation alone.[6]

For example, solely on the basis of the follow-relation we are not able to discover the fact that upon the successive application of LRs 1 and 2, neither LR 1 nor 2 can be applied again. Taking into account the propagation of specifications, the result of the successive application of LR 1 and LR 2 in any order (leading to state q7 or q9) bears the value + on features W and Y. This excludes LRs 1 and 2 as possible followers of that sequence since their in-specifications do not unify with those values. As a result, the arcs $1(q7, q2)$ and $2(q9, q3)$ can be removed to obtain a *reduced* automaton representing global LR interaction.

[6]Note that in case of transitions belonging to a cycle, only those transitions can be removed which are useless at the first visit and after any traversal of the cycle.

## 2.3 Abstract Lexicon Expansion

In the third compilation step the reduced FSA representing global LR interaction is fine-tuned for each LE in the lexicon. The result is a *pruned* FSA. The pruning is done by performing the LR applications corresponding to the transitions in the automaton representing global LR interaction. If the application of a particular LR fails, we know that the corresponding transition can be pruned for the LE under consideration. In case of indirect or direct cycles in the automaton, however, we cannot derive all possible LEs as there may be infinitely many. Even though certain transitions can be pruned even in such cyclic cases, it is possible that certain "redundant" transitions remain in the pruned automaton. However, this is not problematic since the LR application corresponding to such a transition simply fails during processing.

Consider the following base LE:

$$\text{lex_entry}(\begin{bmatrix} \text{A} & b \\ \text{B} & - \\ \text{C} & {}_{t_2}\begin{bmatrix} \text{W} & - \\ \text{X} & - \\ \text{Y} & - \\ \text{Z} & \langle a, b \rangle \end{bmatrix} \end{bmatrix}).$$

Figure 10: A lexical entry

With respect to this base LE we fine-tune the reduced automaton representing global LR interaction (i.e. figure 9 without arcs $1(q7, q2)$ and $2(q9, q3)$) by pruning transitions. We can prune the transitions $\{3(q2, q8), 4(q2, q6), 3(q3, q11), 4(q3, q10), 3(q1, q4), 4(q1, q5)\}$, because the LRs 3 and 4 can not be applied to a (derived) LE which does not have both W and X of value +. As a consequence the states q8, q15, q11, q18, q4 and q12 are no longer reachable and the following transitions can be eliminated as well: $\{3(q8, q8), 4(q8, q15), 3(q11, q11), 4(q11, q18), 3(q4, q4), 4(q4, q12)\}$. We can also eliminate the transitions $\{4(q7, q13), 4(q9, q17)\}$, because the LR 4 requires Z to be of value empty list. Note that the LRs 3 and 4 remain applicable in $q14$ and $q16$.

Furthermore, due to the procedural interpretation of LRs (in contrast to the original declarative intention behind the LRs by the linguist), there can be sequences of LR applications which produce identical entries. To avoid having arcs in the pruned automaton which lead to such identical entries, we use a tabelling method during abstract lexicon expansion which keeps track of the feature structures obtained for each node. If an identity is detected, one of the arcs leading to the corresponding nodes is discarded. In the example, $q7$ and $q9$ are such identical nodes. So we can discard either $2(q2, q7)$ or $1(q3, q9)$ and eliminate the arcs from states which then become unreachable. Choosing to discard $1(q3, q9)$, the pruned automaton for the example LE looks as displayed in figure 11.[7]

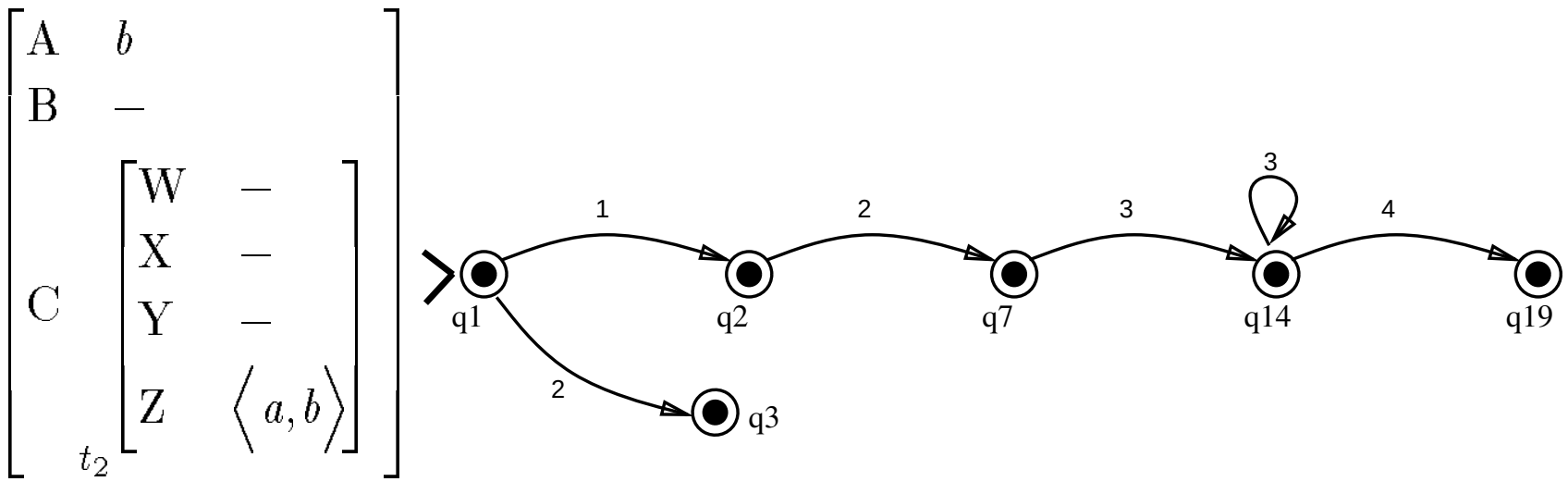

Figure 11: Pruned FSA representing lexical rule interaction for a lexical entry

Note that abstract lexicon expansion does not influence the representation of the LRs themselves. Pruning the FSA representing global LR interaction only involves restricting LR interaction in relation to the LEs in the lexicon.

The fine-tuning of the automaton representing LR interaction results in a FSA for each LE in the lexicon. However, the same automata are obtained for certain groups of LEs and, as shown in the next section, each automaton is translated into definite relations only once. Note that these groups of LEs correspond to the natural classes for which the linguist intended a certain sequence of LR applications to be possible. No additional hand specification is required. So performing abstract expansion for every LE in the lexicon is not as unattractive as it might seem at first sight. Moreover, the alternative computational treatment to expand out the full lexicon at compile time is just as costly and furthermore impossible in case of an infinite lexicon.

## 2.4 Lexical Rule Interaction as Definite Relations

In the fourth compilation step, the FSAs produced in the last step are encoded in definite clauses, so-called, *interaction predicates*. The LEs belonging to a particular natural class all call the interaction predicate encoding the automaton representing LR interaction for that class. Figure 12 shows the extended version of the LE of figure 10.

$$\text{lex_entry}(\boxed{Out})\text{:- q_1}\left(\begin{bmatrix} \text{A} & b \\ \text{B} & - \\ \text{C} & \begin{bmatrix} \text{W} & - \\ \text{X} & - \\ \text{Y} & - \\ \text{Z} & \langle a,b \rangle \end{bmatrix}_{t_2} \end{bmatrix}, \boxed{Out}\right).$$

Figure 12: An *extended* lexical entry

[7]Note that an automaton can be made even more deterministic by unfurling instances of cycles prior to pruning. In our example, unfurling the direct cycle by replacing $3(q14, q14)$ with $\{3(q14, q14'), 3(q14', q14'), 4(q14', q19')\}$ would allow pruning of the cyclic transition $3(q14', q14')$ and the transition $4(q14, q19)$. Note, however, that unfurling of the first $n$ instances of a cycle does not always reduce nondeterminism. Whether unfurling allows pruning of transitions depends on the grammar, namely the LEs and certain properties of LRs occurring in cycles.

The base LE is fed into the first argument of the call to the interaction predicate *q_1*. For each solution to a call to *q_1* the value of $\boxed{Out}$ is a derived LE.

Encoding a FSA as definite relations is rather straightforward. In fact one can view both representations as notational variants of one another.[8] Each transition in the automaton is translated into a definite relation in which the corresponding LR predicate is called, and each final state is encoded by a unit clause. Using an accumulator passing technique (cf. [O'K90]) we ensure that upon execution of a call to the interaction predicate *q_1* a new LE is derived as the result of successive application of a number of LRs. Note that because of the abstract lexicon expansion step discussed in 2.3, we avoid LR applications that are guaranteed to fail and those which produce identical entries.

The interaction predicate encoding the FSA in figure 11 looks as follows:

```
q_1([In],[Out]):-  lex_rule_1([In],[Aux]), q_2([Aux],[Out]).
q_1([In],[Out]):-  lex_rule_2([In],[Aux]), q_3([Aux],[Out]).
q_2([In],[Out]):-  lex_rule_2([In],[Aux]), q_7([Aux],[Out]).
q_7([In],[Out]):-  lex_rule_3([In],[Aux]), q_14([Aux],[Out]).
q_14([In],[Out]):- lex_rule_3([In],[Aux]), q_14([Aux],[Out]).
q_14([In],[Out]):- lex_rule_4([In],[Aux]), q_19([Aux],[Out]).

q_1([In],[In]).  q_2([In],[In]).  q_3([In],[In]).  q_7([In],[In]).  q_14([In],[In]).  q_19([In],[In]).
```

Figure 13: Defining the interaction of lexical rule predicates

# 3 Abstract Lexicon Expansion Revisited

The automata resulting from abstract lexicon expansion group the LEs into natural classes. In case the automata corresponding to two LEs are identical they belong to the same natural class. However, to each LR application, i.e., to each transition in an automaton, corresponds a transfer predicate which can have a large number of defining clauses. Intuitively understood, each defining clause corresponds to a subclass of the class of LEs to which a LR can be applied. In this section we show that abstract lexicon expansion can in many cases be improved such that it directly groups LEs into subclasses (section 3.1). This means that the redundant nondeterminism resulting from multiply defined transfer predicates can be eliminated. In section 3.2 we discuss how this can be done without splitting up the LR predicates.

## 3.1 Reducing Nondeterminism resulting from Property Transfer

In section 2.1 we introduced transfer predicates with several defining clauses to ensure correct property transfer for the different subclasses of LEs to which a LR can be applied. During abstract lexicon expansion, however, when the FSA representing global LR application is pruned with respect to a particular base LE, we know which subclass we are dealing with. For each interaction definition we can therefore check which of the transfer clauses are applicable and discard the non-applicable ones. We thereby eliminate the redundant nondeterminism resulting from multiply defined transfer predicates. This way to proceed corresponds to a program transformation technique referred to as *deletion of clauses with a finitely failed body* in [PP94]. However, deleting non-applicable transfer clauses would force us to have separate definitions of the LR predicates for each LE. We therefore only "keep track" of the transfer possibilities with respect to a specific LE and do not change the transfer predicates. In the next section, the transfer possibilities are included in the encoding by lifting the specifications of the applicable transfer clauses to the level of the interaction predicates called by the specific LEs. This allows us to eliminate the transfer predicates altogether.

[8] For a description of the compiler as a sequence of program transformations, cf. [Min].

## 3.2 Partial Unfolding

The elimination of the transfer predicates is based on Unfold/Fold transformation techniques ([TS84]). The *unfolding* transformation is also referred to as partial execution. Intuitively understood, unfolding comprises the evaluation of a particular literal in the body of a clause at compile time. As a result, the literal can be removed from the body of the clause. When all occurrences of a particular literal in a program are unfolded, its defining clauses can be eliminated from the program. Whereas unfolding can be viewed as a symbolic way of going forward in computation, *folding* constitutes a symbolic step backwards in computation.

Given a LE as in figure 10, we can discard all transfer clauses which presuppose $t_1$ as value of C as discussed in the previous section. To eliminate the transfer predicates completely, we can successively unfold the transfer predicates and the LR predicates with respect to the interaction predicate. However, such a transformation would result in the loss of a representation of the LR predicates which is independent of a particular LE. Since the independent representation of LRs reflects the fact that LRs can be called by various interaction predicates, i.e., that they constitute generalizations over the complete lexicon, it is preferable to eliminate the transfer predicates without losing the independent representation of the LRs. Our compiler therefore performs what can be viewed as "partial" unfolding: it unfolds the transfer predicates directly with respect to the interaction predicates. One can also view this transformation as successive unfolding of the transfer predicates and the LR predicates with respect to the interaction predicates followed by a folding transformation which isolates the original LR predicates. The resulting definite clause encoding of interaction looks as follows:

$$\text{q_1}(\boxed{In}\begin{bmatrix}\text{B} & \boxed{1}\\ \text{C} & {}_{t_2}\!\begin{bmatrix}\text{W} & \boxed{2}\\ \text{Z} & \boxed{3}\end{bmatrix}\end{bmatrix},\boxed{Out})\text{:- lex_rule_1}(\boxed{In},\boxed{Aux}),\ \text{q_2}(\boxed{Aux}\begin{bmatrix}\text{B} & \boxed{1}\\ \text{C} & {}_{t_2}\!\begin{bmatrix}\text{W} & \boxed{2}\\ \text{Z} & \boxed{3}\end{bmatrix}\end{bmatrix},\boxed{Out}).$$

$$\text{q_1}(\boxed{In}\begin{bmatrix}\text{A} & \boxed{1}\\ \text{B} & \boxed{2}\\ \text{C} & {}_{t_2}\!\begin{bmatrix}\text{X} & \boxed{3}\\ \text{Y} & \boxed{4}\\ \text{Z} & \boxed{5}\end{bmatrix}\end{bmatrix},\boxed{Out})\text{:- lex_rule_2}(\boxed{In},\boxed{Aux}),\ \text{q_3}(\boxed{Aux}\begin{bmatrix}\text{A} & \boxed{1}\\ \text{B} & \boxed{2}\\ \text{C} & {}_{t_2}\!\begin{bmatrix}\text{X} & \boxed{3}\\ \text{Y} & \boxed{4}\\ \text{Z} & \boxed{5}\end{bmatrix}\end{bmatrix},\boxed{Out}).$$

$$\text{q_2}(\boxed{In}\begin{bmatrix} \text{A} & \boxed{1} \\ \text{B} & \boxed{2} \\ \text{C} & {}_{t_2}\begin{bmatrix} \text{X} & \boxed{3} \\ \text{Y} & \boxed{4} \\ \text{Z} & \boxed{5} \end{bmatrix} \end{bmatrix}, \boxed{Out})\text{:- lex_rule_2}(\boxed{In}, \boxed{Aux}),\ \text{q_7}(\boxed{Aux}\begin{bmatrix} \text{A} & \boxed{1} \\ \text{B} & \boxed{2} \\ \text{C} & {}_{t_2}\begin{bmatrix} \text{X} & \boxed{3} \\ \text{Y} & \boxed{4} \\ \text{Z} & \boxed{5} \end{bmatrix} \end{bmatrix}, \boxed{Out}).$$

$$\text{q_7}(\boxed{In}\begin{bmatrix} \text{A} & \boxed{1} \\ \text{B} & \boxed{2} \\ \text{C} & {}_{t_2}\begin{bmatrix} \text{W} & \boxed{3} \\ \text{X} & \boxed{4} \end{bmatrix} \end{bmatrix}, \boxed{Out})\text{:- lex_rule_3}(\boxed{In}, \boxed{Aux}),\ \text{q_14}(\boxed{Aux}\begin{bmatrix} \text{A} & \boxed{1} \\ \text{B} & \boxed{2} \\ \text{C} & {}_{t_2}\begin{bmatrix} \text{W} & \boxed{3} \\ \text{X} & \boxed{4} \end{bmatrix} \end{bmatrix}, \boxed{Out}).$$

$$\text{q_14}(\boxed{In}\begin{bmatrix} \text{A} & \boxed{1} \\ \text{B} & \boxed{2} \\ \text{C} & {}_{t_2}\begin{bmatrix} \text{W} & \boxed{3} \\ \text{X} & \boxed{4} \end{bmatrix} \end{bmatrix}, \boxed{Out})\text{:- lex_rule_3}(\boxed{In}, \boxed{Aux}),\ \text{q_14}(\boxed{Aux}\begin{bmatrix} \text{A} & \boxed{1} \\ \text{B} & \boxed{2} \\ \text{C} & {}_{t_2}\begin{bmatrix} \text{W} & \boxed{3} \\ \text{X} & \boxed{4} \end{bmatrix} \end{bmatrix}, \boxed{Out}).$$

$$\text{q_14}(\boxed{In}\begin{bmatrix} \text{A} & \boxed{1} \\ \text{C} & {}_{t_2}\begin{bmatrix} \text{W} & \boxed{2} \\ \text{Y} & \boxed{3} \\ \text{Z} & \boxed{4} \end{bmatrix} \end{bmatrix}, \boxed{Out})\text{:- lex_rule_4}(\boxed{In}, \boxed{Aux}),\ \text{q_19}(\boxed{Aux}\begin{bmatrix} \text{A} & \boxed{1} \\ \text{C} & {}_{t_2}\begin{bmatrix} \text{W} & \boxed{2} \\ \text{Y} & \boxed{3} \\ \text{Z} & \boxed{4} \end{bmatrix} \end{bmatrix}, \boxed{Out}).$$

$$\text{q_1}(\boxed{In},\boxed{In}). \quad \text{q_2}(\boxed{In},\boxed{In}). \quad \text{q_3}(\boxed{In},\boxed{In}). \quad \text{q_7}(\boxed{In},\boxed{In}). \quad \text{q_14}(\boxed{In},\boxed{In}). \quad \text{q_19}(\boxed{In},\boxed{In}).$$

Figure 14: Unfolding the transfer predicates for the LE of figure 10 with respect to the interaction predicate of figure 13

The LR predicates called by these interaction predicates are defined as in figures 4 and 7 except for the fact that the transfer predicates are no longer called.

# 4 On the Fly Application of Lexical Rules

We want our compiler to produce an encoding of LRs which allows us to execute LRs "on the fly", i.e. at some time *after* lexical lookup. The advantage of such delayed evaluation is that while the execution of the interaction predicate is delayed, more constraints on the LE are collected in processing. When the interaction predicate is finally called, many of its possible solutions simply fail. The search tree which would have resulted from pursuing these possibilities at the beginning of processing does not have to be explored.

As it stands, our encoding of LRs and their application as covariation in LEs does not yet support the application of LRs on the fly. With respect to processing, the extended LE of figure 12 is problematic because before execution of the call to *q_1* it is unknown which information of the base LE ends up in a derived LE. One is therefore forced to execute the call to *q_1* directly when the LE is used during processing. Otherwise there is no information available to restrict the search space of a generation or parsing process. In the following we show how the additional specifications needed on the extended LE to guide processing can be automatically deduced.

The intuitive idea is to lift the information which is ensured after all sequences of LR applications which are possible for a particular base LE into the extended LE. Note that this is not an unfolding step. Unfolding the interaction predicates with respect to the LEs would lead to an increase of the number of LEs in effect comparable to off-line lexicon expansion. Instead, what we do is factor out the information which is common to all definitions of the called interaction predicate through computing the generalization of these definitions. We then use the obtained generalization to enrich the extended LE.

The generalization can contain much valuable information because it is usually the case that LEs resulting from LR application only differ in very few specifications compared to the number of specifications in a base LE. Most of the specifications of a LE are assumed to be passed unchanged via automatic property transfer. After lifting this information into the extended LE, the out-argument in many cases contains enough information to permit a delayed execution of the interaction predicate.

To illustrate this final step, we show how a LE suitable for on the fly application is obtained. Since the running example of this paper was kept small for expository reasons by only including features that do get changed by one of the LRs (which violates the empirical observation discussed above), the full set of LRs will not provide a good example. Let us therefore assume that only the LRs 1 and 2 of figure 6 are given. We then only obtain seven of the clauses of figures 13 and 14: those calling *lex_rule_1* or *lex_rule_2*, as well as the unit clauses for *q_1*, *q_2*, *q_3*, and *q_7*.

We lift the information unchanged by this interaction predicate into the extended LE of figure 12 using a technique that is similar to the technique used in [MGG95] for off-line optimization of phrase structure rules in typed feature structure grammars. We evaluate the interaction predicate off-line in a bottom-up fashion. However, when there is more than one defining clause with which a right-hand side literal unifies, we do not pick one of them, but consider all clauses and unify the generalization of their head literals with the right-hand side literal. Once we reach the top-level interaction predicate we unify the generalization of the head literals of its defining clauses with the call to the interaction predicate in the extended LE. As a result, all information left unchanged by the evaluation of the call to the interaction predicate is lifted up into the LE and becomes available upon lexical lookup. Applying this technique to the extended LE of figure 12 yields the following result:

$$\text{lex_entry}(\boxed{Out}\begin{bmatrix} \text{A} & b \\ \text{B} & \boxed{2} \\ \text{C} & {}_{t_2}\!\begin{bmatrix} \text{Z} & \boxed{3} \end{bmatrix} \end{bmatrix}) \text{:- q_1}(\begin{bmatrix} \text{A} & b \\ \text{B} & \boxed{2}\ - \\ \text{C} & {}_{t_2}\!\begin{bmatrix} \text{W} & - \\ \text{X} & - \\ \text{Y} & - \\ \text{Z} & \boxed{3}\left\langle a, b \right\rangle \end{bmatrix} \end{bmatrix}, \boxed{Out}).$$

Figure 15: An entry suitable for on the fly application (LR 1 and 2 only)

Even though we see on the fly application as a prerequisite of a computational treatment of LRs, it is important to note that delayed evaluation of LR application is not always profitable. For example, underspecification of the head of a construction can lead to massive nondeterminism or even nontermination when not enough restricting information is available to generate its complements. Criteria to determine whether or not to delay the evaluation of a LR are needed. [vNB94] suggest to use goal-freezing to decide whether to delay the evaluation of LR application. This necessitates the procedural annotation of otherwise declarative specifications. The linguist has to specify restrictions on the instantiation status of a goal which need be fulfilled before it can be executed. Thus the approach presupposes that the linguist possesses substantial computational expertise. Furthermore, the computational bookkeeping necessary for the freezing mechanism is very expensive. We therefore think that it is preferable to deal with these kind of control problems in a static fashion along the lines of [MGG95] and [MGH]. They describe the use of a dataflow analysis for an off-line grammar optimization which determines automatically when a particular goal can best be executed.

# 5 Concluding Remarks

We presented a new computational treatment of HPSG LRs by describing a compiler which translates a set of LRs as specified by a linguist into definite relations which are used to constrain LEs. We determine LR interaction and represent it by a FSA. The automaton enables us to avoid the derivation of duplicate LEs and allows us to encode LR interaction without actually having to execute a possibly infinite number of calls to LRs. By means of abstract lexicon expansion the finite state automaton is refined in order to avoid LR applications that are guaranteed to fail. The refined automaton is encoded in definite relations without losing any of the generalizations captured by LRs. Finally, adapting the LEs, we make it possible to apply LRs on the fly.

Building on the work described in this paper, we used the encoding of LRs and their application as covariation in LEs in the Troll system ([GK94]) for a grammar implementing a complex HPSG theory covering the so-called aux-flip phenomenon and partial-VP topicalization in the three clause types of German ([Meu94]). Given a set of five LRs, the proposed LR encoding lead to a lexicon which is about fifty percent smaller than the one obtained from off-line expansion (from which duplicate entries were already removed).

# References

[Car92] B. Carpenter. ALE – the attribute logic engine, user's guide. CMU-LCL report 92-1, Carnegie Mellon University, 1992.

[Fli87] D. Flickinger. *Lexical rules in the hierarchical lexicon.* PhD thesis, Stanford University, 1987.

[FPW85] D. Flickinger, C. Pollard, and T. Wasow. A computational semantics for natural language. In *Proc. of the 23rd Annual Meeting of the* ACL, 1985.

[Gei94] S. Geißler. Lexikalische Regeln in der IBM-Basisgrammatik. Verbmobil report 20, Institute for logic and linguistics, IBM Heidelberg, 1994.

[GK94] D. Gerdemann and P. J. King. The correct and efficient implementation of appropriateness specifications for typed feature structures. In *Proc. of* COLING 94, Kyoto, 1994.

[GM95] T. Götz and W. D. Meurers. Compiling HPSG type constraints into definite clause programs. In *Proc. of the 33rd Annual Meeting of the* ACL, 1995.

[Kin89] P. J. King. *A logical formalism for Head-Driven Phrase Structure Grammar.* PhD thesis, University of Manchester, 1989.

[Kin94] P. J. King. An expanded logical formalism for Head-Driven Phrase Structure Grammar. SFB 340 report 59, Universität Tübingen, 1994.

[Meu94] W. D. Meurers. On implementing an HPSG theory – Aspects of the logical architecture, the formalization, and the implementation of head-driven phrase structure grammars. In E. W. Hinrichs, W. D. Meurers, and T. Nakazawa (eds.), *Partial-*VP *and split-*NP *topicalization in German – An* HPSG *analysis and its implementation.* SFB 340 report 58. Universität Tübingen, 1994.

[Meu] W. D. Meurers. A semantics for lexical rules as used in HPSG. SFB 340, Universität Tübingen, in preparation.

[MGG95] G. Minnen, D. Gerdemann, and T. Götz. Off-line optimization for Earley-style HPSG processing. In *Proc. of the 7th Conference of the* EACL, Dublin, 1995.

[MGH] G. Minnen, D. Gerdemann, and E. Hinrichs. Direct automated inversion of logic grammars. *New Generation Computing*, to appear. An earlier version appeared in Y. Matsomoto (ed.) *Proc. of the 4th International Workshop on Natural Language Understanding and Logic Programming*, 1993.

[Min] G. Minnen. *Reversible grammar: Off-line compilation techniques for analysis and synthesis of natural language.* PhD thesis, in preparation.

[O'K90] R. O'Keefe. *The craft of prolog.* MIT Press, Cambridge, Mass, 1990.

[Pol93] C. Pollard. Lexical rules as metadescriptions. Handout for "Between Syntax, Semantics, and Logic" Conference, Universität Stuttgart, Oct. 5, 1993.

[PP94] A. Pettorossi and M. Proietti. Transformations of logic programs: Foundations and techniques. *Journal of Logic Programming*, 1994.

[PS94] C. Pollard and I. A. Sag. *Head-Driven Phrase Structure Grammar.* University of Chicago Press, Chicago, 1994.

[TS84] Y. Tamaki and T. Sato. Unfold/fold transformation of logic programs. In S. Tarnlund (ed.), *Proc. of the 2nd Int. Conference on Logic Programming*, Uppsala, 1984.

[vNB94] G. van Noord and G. Bouma. The scope of adjuncts and the processing of lexical rules. In *Proc. of* COLING 94, Kyoto, 1994.